%% file: paper.tex
\documentclass[letterpaper]{article} 
\usepackage{aaai2026}  
\usepackage{times}  
\usepackage{helvet}  
\usepackage{courier}  
\usepackage[hyphens]{url}  
\usepackage{graphicx} 
\usepackage{natbib}  
\usepackage{caption} 
\usepackage{algorithm}
\usepackage{algorithmic}
\usepackage{listings}
\usepackage{babel}
\usepackage[inline]{enumitem}
\usepackage[fixed]{fontawesome6}
\usepackage{fontspec}
\usepackage{pgfplots}
\usepackage{tikz}
\usepackage{xcolor}
\usepackage{listings}
\usepackage{fancyvrb}
\usetikzlibrary{
    arrows,
    automata,
    backgrounds,
    calc,
    decorations.shapes,
    decorations.text,
    fit,
    mindmap,
    patterns,
    petri,
    positioning,
    shadows,
    shapes,
    shapes.gates.logic.US,
    trees
}
\definecolor{tol-hc-red}{HTML}{BB5566} 
\definecolor{tol-hc-blue}{HTML}{004488} 
\definecolor{tol-hc-lightblue}{HTML}{2291ff} 
\definecolor{tol-hc-yellow}{HTML}{DDAA33} 
\definecolor{tol-legend}{HTML}{666666}
\definecolor{tol-hc-green}{HTML}{2E6F40}

\usepackage[tickmarkheight=0.1cm, textsize=tiny, textwidth=1.4cm]{todonotes}

\usepackage[acronym]{glossaries} 
\glsdisablehyper 
\loadglsentries{acronyms.tex}

\usepackage{minted}

\usepackage{dirtytalk}
\usepackage{amsfonts}

\newcommand{\fungai}{\texttt{FungAI}}
\newcommand{\wfterm}[1]{\textsc{\sffamily #1}}
\newcommand{\wfiterm}[2]{\faIcon{#1}\,\wfterm{#2}}
\newcommand{\threatProtect}[1]{\textcolor{tol-hc-blue}{#1}}
\newcommand{\threatDetect}[1]{\textcolor{tol-hc-lightblue}{#1}}
\newcommand{\threatFail}[1]{\textcolor{tol-hc-red}{#1}}
\newcommand{\threatIconProtect}{\textcolor{tol-hc-blue}{\faIcon{shield-halved}}}
\newcommand{\threatIconDetect}{\textcolor{tol-hc-lightblue}{\faIcon{bell}}}
\newcommand{\threatIconFail}{\textcolor{tol-hc-red}{\faIcon{user-secret}}}

\usepackage{newfloat}
\usepackage{listings}
\DeclareCaptionStyle{ruled}{labelfont=normalfont,labelsep=colon,strut=off} 
\floatstyle{ruled}
\newfloat{listing}{tb}{lst}{}
\floatname{listing}{Listing}

\title{A Workflow for Full Traceability of AI Decisions}

\author {
    Julius Wenzel\textsuperscript{\rm 1},
    Syeda Umaima Alam\textsuperscript{\rm 2},\\
    Andreas Schmidt\textsuperscript{\rm 2},
    Hanwei Zhang\textsuperscript{\rm 2},
    Holger Hermanns\textsuperscript{\rm 2}
}
\affiliations {
    \textsuperscript{\rm 1}Technische Universität Dresden,
    \textsuperscript{\rm 2}Saarland University\\
    julius.wenzel@tu-dresden.de, syal00002@stud.uni-saarland.de,{schmidt,zhang,hermanns}@depend.uni-saarland.de
}

\usepackage{bibentry}

\begin{document}

\newcommand{\takeaway}[1]{\begin{center}
\colorbox{white!90!blue}{\begin{minipage}{0.95\columnwidth} \emph{#1}\end{minipage}}
\end{center}} 

\maketitle

\input{sections/00-abstract.tex}


\input{sections/00-availability.tex}
\input{sections/01-intro.tex}
\input{sections/02-background.tex}
\input{sections/03-ai-workflow.tex}
\input{sections/04-threats.tex}

\input{sections/05-dbom-format.tex}
\input{sections/06-depend-dbom.tex}

\input{sections/07-use-case.tex}

\input{sections/07-eval.tex}

\input{sections/08-ecosystem.tex}
\input{sections/09-conclusion.tex}


\bibliography{aaai2026}

\input{sections/11-supp}

\end{document}

%% file: acronyms.tex
\newacronym{cas}{CAS}{Configuration and Attestation Service}
\newacronym{dbom}{DBOM}{Decision Bill of Material}
\newacronym{dsse}{DSSE}{Dead Simple Signing Envelope}
\newacronym{hci}{HCI}{Human-Computer Interaction}
\newacronym{ibom}{DBOM\babelhyphen{nobreak}I}{Inference Decision Bill of Material}
\newacronym{ood}{OOD}{Out-of-Distribution}
\newacronym{sbom}{SBOM}{Software Bill of Material}
\newacronym{sgx}{SGX}{Software Guard Extensions}
\newacronym{tbom}{DBOM\babelhyphen{nobreak}T}{Training Decision Bill of Material}
\newacronym{tee}{TEE}{Trusted Execution Environment}

%% file: sections/00-abstract.tex
\begin{abstract}
    An ever increasing number of high-stake decisions are made or assisted by automated systems employing brittle artificial intelligence technology.
    There is a substantial risk that some of these decision induce harm to people, by infringing their well-being or their fundamental human rights.
    The state-of-the-art in AI systems makes little effort with respect to appropriate documentation of the decision process.
    This obstructs the ability to trace what went into a decision, which in turn is a prerequisite to any attempt of reconstructing a responsibility chain.
    Specifically, such traceability is linked to a documentation that will stand up in court when determining the cause of some AI-based decision that inadvertently or intentionally violates the law.
    
    This paper takes a radical, yet practical, approach to this problem, by enforcing the documentation of each and every component that goes into the training or inference of an automated decision.
    As such, it presents the first running workflow supporting the generation of tamper-proof, verifiable and exhaustive traces of AI decisions. In doing so, we expand the \gls{dbom} concept \cite{wenzel2024traceability} into an effective running workflow leveraging confidential computing technology.
    We demonstrate the inner workings of the workflow in the development of an app to tell poisonous and edible mushrooms apart, meant as a playful example of high-stake decision support.
    \glsresetall 
\end{abstract}

%% file: sections/00-availability.tex

%% file: sections/01-intro.tex
\section{Introduction}
\label{sec:introduction}

As AI models are increasingly integrated into our daily lifes, ensuring their trustworthiness has become a pressing concern.
The widespread deployment of AI systems, particularly in sensitive and high-stakes domains, demands rigorous scrutiny.
The European Union's AI Act~\cite{AIAct} underscores this urgency by mandating that AI systems classified as \emph{high-risk} must address key issues such as transparency, interpretability, cybersecurity, and data privacy.
This echoes a growing global recognition that trustworthy AI is not optional but essential for safe and ethical deployment.
However, ensuring trustworthiness is far from straightforward.
The opacity and complexity of modern AI models pose significant challenges for auditing and compliance, especially in high-risk scenarios.
Traditional interpretability methods fall short in providing the end-to-end transparency needed for robust oversight.

\paragraph{Bill of Materials.} 
We propose a 
holistic and practical approach to this problem, enabled by treating AI models as complex software systems.
By reviewing the entire implementation pipeline from data preparation, server infrastructure, and training procedures to deployment and inference, we establish tamper-proof traceability at and across each stage.
Building on the concept of \gls{dbom}~\cite{wenzel2024traceability}, we propose and implement a practical approach for documenting exhaustively the components of AI system development and decision-making processes. We use cryptograpic technology to make the entire documentation tamper-proof and traceable, and this is supported by  \gls{dbom}-inspection technology we develop. 

\paragraph{Use Case.}
As a running example when explaining and experimenting with the \gls{dbom} workflow we present 
\fungai{}, a practical mobile phone application that can help determine whether certain mushrooms~(funghi) are poisonous, based on structured data and images.
Such an application could support foraging safety, educational tools, or assist in biodiversity research by providing quick, automated assessments of mushroom toxicity.
Wrong classifications may come with high-stake consequences if blindly followed.
In this respect, the app shares crucial characteristics with more serious high-stake applications, such as skin cancer recognition apps or other medical image classification systems.

\paragraph{Contributions.} The paper contributes the following:
\begin{itemize}
    \item We present a fully functional \gls{dbom} workflow, enabling tamper-proof, verifiable and exhaustive traces of AI decisions.
    \item We discuss threats to the dependable functioning of a generic AI system, and in how far \glspl{dbom} protect against these threats.
    \item We apply the approach to the \fungai{} use case. 
   \item We discuss first empirical evaluations together with first components of an ecosystem of \gls{dbom}-based tools.
\end{itemize}

\paragraph{Organization of the Paper.}
We continue this paper by reviewing the scientific context of the work. 
A characterization of a generic AI system is the base for discussing  threats to the dependability of AI decision systems, and the role of \glspl{dbom} in protecting against or detecting these threats.
We then turn to explaining how \glspl{dbom} are generated dependably and how they serve in the concrete \fungai{} use case. Details of the \gls{dbom} implementation are discussed and empirically evaluated, before  we give an outlook on a \gls{dbom} ecosystem and conclude the paper.

%% file: sections/02-background.tex
\section{Background \& Related Work}
\label{sec:background}

\paragraph{AI Workflows \& Accountability.}

There is a body of work considering AI workflows in their entirety and attempting to improve the transparency, accountability, and traceability of the entire process. Efforts such as Datasheets~\cite{gebru2021datasheets} and Model Cards~\cite{mitchell2019model} propose standardized, structured documentation. AIQPROV~\cite{nakagawa2022provenance} extends standard provenance recodes by incorporating human activities. However, they rely on manual, post hoc templates, often cover only parts of the AI workflow, and limit scalability and cross-platform comparability. Tools such as MLflow~\cite{zaharia2018accelerating}, MLflow2Prov~\cite{schlegel2023mlflow2prov}, DLProv~\cite{pina2024dlprov}, OpenLineage~\cite{openlineage}, Vertex AI~\cite{vertexai}, and AIPassport~\cite{kalokyri2025ai} enable automated metadata generation. However, MLflow, MLflow2Prov, DLProv, and OpenLineage cover only parts of the AI lifecycle; Vertex AI is not open-source, and AIPassport is domain-specific to healthcare. While these tools support traceability, they are not guaranteed to be tamper-proof.
Another research direction explores auditing methods, such as checklists, frameworks, and engineering practices, to improve traceability and accountability across the AI lifecycle~\cite{raji2020closing}. Human-computer interaction research further examines how workflows, interfaces, and organizational processes shape accountability in practice~\cite{metcalf2021algorithmic,he2025fine}.
Distinct from prior efforts, \gls{dbom}~\cite{wenzel2024traceability} introduces a conceptual framework for traceable decision bills of materials combined with confidential computing to enable auditing, security assurance, and human oversight. It also offers a foundation for embedding AI alignment values. Building on this vision, our work presents a practical, technically implementable solution that advances \gls{dbom} from concept to application.

\paragraph{Confidential Computing.}
Confidential Computing is a technique that allows to shield program execution---even from powerful attackers.
It generally relies on hardware features, i.e. protection mechanisms that are carried out inside the hardware that cannot be influenced (setting aside side-channel attacks).
Initially, confidential computing was a CPU feature: Starting with Intel (\gls{sgx}) in 2015, other processor technologies, such as AMD-SEV and Intel TDX, have emerged.
Confidential Computing is closely related to the notion of a \gls{tee}, a secure space separated from the rest of the computer system.
Being hardware-related, Confidential Computing applications were often difficult to adapt for developers, but frameworks such as SCONE \cite{arnautov2016scone} or Graphene \cite{tsai2017graphene} have simplified this process.
Nowadays, there is an increasing number of Confidential Computing applications in use \cite{sgxsurvey} and applications in sensitive domains, such as finance \cite{georg2023nautilus} and health showing that the technology has achieved maturity.

%% file: sections/03-ai-workflow.tex
\section{A Generic AI System and its Use}
\label{sec:ai-system}

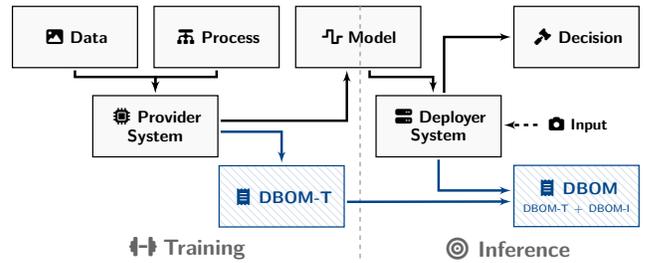
\begin{figure}[t]
      \resizebox{1.0\columnwidth
      }{!}{\input{figures/ai-workflow.tex}}
      \caption{
            A generic AI system can be separated in components and information flows in the direction of the arrows.
            Our contributions are marked in \textcolor{tol-hc-blue}{blue}.
      }
      \label{fig:abstract-workflow}
\end{figure}

We start off by characterizing a generic AI system~(cf. Figure~\ref{fig:abstract-workflow})---where the \gls{dbom} will integrate tightly in many places.
On a high-level, we consider it natural to distinguish between the
\textcolor{tol-legend}{\textbf{\wfiterm{dumbbell}{Training}}} and
\textcolor{tol-legend}{\textbf{\wfiterm{bullseye}{Inference}}}.
In detail, we find the following components:
\begin{description}
      \item[\wfiterm{image}{Data}]
            These are labelled or unlabeled samples~(e.g. mushroom images) as input to the training process.
      \item[\wfiterm{chart-diagram}{Process}]
            This encompasses the design of a suitable model, pre-processing of data, execution of training algorithms, and evaluation of model performance---potentially done repetitively.
            For our purposes, this highly aggregated and abstract view is sufficient.
      \item[\wfiterm{microchip}{Provider System}]
            The provider of an AI system employs hardware and runs training processes on data---producing a trained model,  together with a \gls{tbom}.
      \item[\wfiterm{wave-square}{Model}]
            This is 
            the end-product of training that is later used in inference.
      \item[\wfiterm{server}{Deployer System}]
            The deployer is the entity that uses an AI system under its authority. It executes the inference based on the model and applies the trained model to new (e.g.~mushroom) input samples, yielding a decision and an \gls{ibom}.
      \item[\wfiterm{gavel}{Decision}]
            The decision is the result of the inference and is presented to the user.
\end{description}
In our approach, a trained model is accompanied by a training \gls{dbom}~(\textcolor{tol-hc-blue}{\wfiterm{receipt}{\acrshort{tbom}}}).
Once a decision is made, an inference \gls{dbom} ~(\textcolor{tol-hc-blue}{\wfiterm{receipt}{\acrshort{ibom}}}) is produced that also stores a reference to \gls{tbom}.

%% file: figures/ai-workflow.tex
\begin{tikzpicture}[font=\sffamily]

    \definecolor{bg}{HTML}{F9F9F9}
    \tikzset{
        hatched/.style={pattern=north west lines,pattern color=#1},
        hatched/.default=black
    }
    \tikzstyle{b} = [rectangle, draw=black, fill=bg, node distance=1.75em and -3.0em, text=black, text width=7.5em, text centered, minimum height=4.0em, thick, font={\large\sffamily}]
    \tikzstyle{mod_b} = [b, text=tol-hc-blue, draw=tol-hc-blue, hatched=tol-hc-blue!15!white]
    \tikzstyle{input_b} = [b, dashed]
    \tikzstyle{l} = [draw=black, text=black, -latex', ultra thick, shorten <=0.5mm, shorten >=0.5mm]
    \tikzstyle{mod_l} = [l,draw=tol-hc-blue]
    \tikzstyle{legend} = [font={\small\sffamily\bfseries},text=tol-legend]
    \tikzstyle{ll} = [align=center, font={\small\sffamily}]
    \tikzstyle{mod_ll} = [ll, below, text=tol-hc-blue]
    \tikzstyle{threat} = [node distance=-1.25em and -1.25em]
    \tikzstyle{threat_s} = [threat]
    \tikzstyle{threat_f} = [threat]

    \def\height{16.5em}

    \node [b] (provider) {
        \textbf{\wfiterm{microchip}{\large Provider\\System}}
    };
    \node [b, above left= of provider] (training_data) {
        \textbf{\wfiterm{image}{Data}}
    };
    \node [b, right=1em of training_data] (training_process) {
        \textbf{\wfiterm{chart-diagram}{Process}}
    };

    \node [b, right=1em of training_process] (trained_model) {
        \textbf{\wfiterm{wave-square}{Model}}
    };
    \node [mod_b, below right=0.5em and 0em of provider] (DBOM-T) {
        \textbf{\wfiterm{receipt}{DBOM-T}}
    };
    \coordinate (l0) at ([yshift=1em]training_data.north west);

    \node [b, below right= of trained_model] (user) {\textbf{\wfiterm{server}{Deployer\\System}}};
   
    \node [b, right=6em of trained_model] (decision) {\textbf{\wfiterm{gavel}{Decision}}};
    
    
    \node [below= of decision] (input)
    {\textbf{\wfiterm{camera}{Input}}};

    \node [mod_b, below= of input] (dbom) {
        \textbf{\wfiterm{receipt}{DBOM}}\\
        \scriptsize DBOM-T + DBOM-I
    };

    \coordinate (l1) at ([xshift=-4.0em,yshift=0em]trained_model.north east);
    \coordinate (l2) at ([xshift=0.75em,yshift=1em]decision.north east);
    \draw[black!40,thick,dashed] (l1) -- ++(0,-\height);

    \node [legend, below=\height, yshift=1.25em] (lablTrain) at ($(l0)!.5!(l1)$) {\LARGE\textsc{\wfiterm{dumbbell}{Training}}};
    \node [legend, below=\height, yshift=1.25em] (lablInfer) at ($(l1)!.5!(l2)$) {\LARGE\textsc{\wfiterm{bullseye}{Inference}}};

    \draw [l] (training_data.south) -- ++(0,-0.2) -| (provider.north);
    \draw [l] (training_process.south) -- ++(0,-0.2) -| (provider.north);
    \draw [l] (provider.5) -| (trained_model.250);
    \draw [mod_l] (provider.355) -| (DBOM-T.north);
    \draw [l] (trained_model.290) -- ++(0,-0.2) -| (user.100);
    \draw [mod_l] (user.south) |- (dbom.175);
    \draw [mod_l] (DBOM-T.355) -- (dbom.185);
    \draw [l] (user.80) |- (decision.west);
    \draw [l, dashed] (input.west) -- (user.1);

    \node [threat_f, above right= of training_data] {};
    \node [threat_s, above right= of training_process] {};
    \node [threat_s, above right= of user] {};

    \node [threat_s, above right= of trained_model] {};
    \node [threat_s, above right= of provider] {};

\end{tikzpicture}

%% file: sections/04-threats.tex
\section{Dependable AI Decision Systems}

According to regulations such as the EU AI Act~\cite{AIAct}, a reliable AI decision system must provide well-documented, traceable processes and ensure security against potential threats.
Our approach delivers traceable, tamper-proof documentation in the form of a bill of materials.
Under normal conditions without threats, a \gls{dbom} guarantees transparency, traceability, and accountability through distinct components: \gls{tbom}, which documents and enables auditing of the training process (as illustrated in Figure~\ref{fig:abstract-workflow}), and \gls{ibom}, which documents the inference process to support explanations regarding individual decisions.

In the presence of threats, \glspl{dbom} provide
full protection~(\threatIconProtect{}) or only detection~(\threatIconDetect{}).
For some threats, however, our approach is unable to help overcoming them~(\threatIconFail{}).
Below, we outline specific threats to a generic system and how they can (or cannot) be handled.
The component icon resembles where the focus of the threat lies (multiple components can be threatened at the same time).

\newcommand{\threatHeading}[4]{#1{\wfiterm{#3}{}\hspace{-0.5em}{\tiny{#2}} #4}}

\subsection{Threats during \textcolor{tol-legend}{\textbf{\wfiterm{dumbbell}{Training}}}}

\begin{description}
           \item[\threatHeading{\threatProtect}{\threatIconProtect{}}{wave-square}{Model Manipulation (at rest)}]
            refers to attempts to bypass the training and influence decisions directly via the Model.
            Since we protect the integrity our data-at-rest with cryptographic signatures, we can rule out this attack vector.
      \item[\threatHeading{\threatProtect}{\threatIconProtect{}}{microchip}{Data-in-use Manipulation}]
            may occur when an (advanced) attacker attempts to alter the live memory during the training process.
            Our use of \glspl{tee} effectively protects against such attacks by securing data while it is being processed. \item[\threatHeading{\threatDetect}{\threatIconDetect{}}{image}{Training Data Tampering}]
            includes \emph{poisoning and backdoor attacks}, which aim to corrupt model training by altering the training data.
            \emph{Poisoning attacks} subtly modify inputs to degrade model performance, while \emph{backdoor attacks} embed triggers that cause the model to behave maliciously only when activated.
            Although our approach cannot directly detect data manipulation, it facilitates forensic investigations by precisely documenting the datasets used and thus enabling traceability of problematic data points.
      \item[\threatHeading{\threatDetect}{\threatIconDetect{}}{chart-diagram}{Training Process Poisoning}]
            targets the integrity of the training pipeline by compromising the environment, altering training algorithms, or exploiting system vulnerabilities.
            The \gls{dbom} documents the environment and the algorithms used in training.
            It cannot prevent an attack on this part of the training pipeline, but will expose it, as the attacker cannot prevent the alterations from being documented. 
\end{description}

\subsection{Threats during \textcolor{tol-legend}{\textbf{\wfiterm{bullseye}{Inference}}}}
\begin{description}
      \item[\threatHeading{\threatProtect}{\threatIconProtect{}}{server}{Tampered Inference System}]
            is an attack on the inference process's living memory that is prevented again by the usage of \glspl{tee}.
      \item[\threatHeading{\threatProtect}{\threatIconProtect{}}{gavel}{Decision Manipulation during Delivery}]
            would mean replacing the actual AI decision by one under the control of the attacker.
            The \gls{dbom} is tied to the decision with a signature and can be verified, thus protecting the decision against tampering.
      \item[\threatHeading{\threatDetect}{\threatIconDetect{}}{wave-square}{Illegitimate Inputs}]
            include adversarial examples and \gls{ood} data.
            Adversarial inputs involve subtle perturbations that mislead the model, while \gls{ood} inputs occurr when inference data is not adequately represented in the training set. Although \gls{dbom} cannot prevent such errors, it enables retrospective analysis by identifying and tracing the relevant training data.
      \item[\threatHeading{\threatFail}{\threatIconFail{}}{wave-square}{Information Leakage}]
            including model extraction, membership inference, and model inversion, aim to reconstruct model parameters, reveal training data membership, or infer sensitive attributes learned by the model.
            It is not possible to expose all kinds of information leakage in the \gls{dbom} and we cannot prevent a leakage with our current approach.
            Thus, the \gls{dbom} does not address this threat.
\end{description}

%% file: sections/05-dbom-format.tex
\section{Traceability using the \Gls{dbom} Format}
\label{sec:details}

Our approach introduces two core artifacts to enable full traceability of AI decisions:

\begin{description}
    \item[\faIcon{dumbbell} \gls{tbom}]
          captures all relevant information about the training process:
          dataset summaries, preprocessing steps, hyperparameters, cross-validation metrics, as well as hardware and software versions used.
          Model parameters and evaluation results are also documented in the \gls{tbom}.
          This artifact enables reproducibility and auditing of the training workflow.

    \item[\faIcon{bullseye} \gls{ibom}]
          details the inference process for each decision:
          raw input features, encoding strategies, model predictions,  decision logic, as well as hardware and software versions used.
          Additionally, it records timestamps and cryptographic signatures to ensure integrity and authenticity.
\end{description}

\paragraph{\Gls{tbom} Generation.}

Training produces a \gls{tbom} that summarises the information mentioned above. 
The process involves loading and preprocessing the dataset, training the classifier, evaluating through cross-validation, and finally generating a \gls{tbom} record cryptographically signed via \gls{dsse}.

\paragraph{\Gls{tbom} Format.}
The \gls{tbom} is a structured JSON file containing the following key sections:

\begin{description}
    \item[Project Metadata:]
          Documents the high-level purpose of the task and versioning information.
    \item[Data Summary:]
          Provides a complete overview of the data used, including, e.g., total sample counts or class distributions.
          For full reproducibility, this section also stores the exact indices used for the main data splits.
    \item[Model Architecture:]
          A component-wise breakdown of the model.
    \item[Training Methodology:]
          Details the evaluation approach and lists all hyperparameters used, such as learning rate, batch size, optimizer, and epochs per fold. It also specifies that a final model is trained on the full non-test dataset.
    \item[Performance Metrics:]
          A comprehensive report of the model's performance.
          This includes detailed cross-validation statistics (mean accuracy, standard deviation, and per-fold results) as well as the final, unbiased performance metrics (accuracy, sensitivity, specificity, etc.) on the hold-out test set.
    \item[Environment and Dependencies:]
          A manifest of the computational environment, including the hardware (e.g., CPU/GPU), Python version, and key library versions (e.g., PyTorch, scikit-learn).
    \item[Output Artifacts:]
          Contains pointers to the files with the saved final model weights and the \gls{tbom} file itself.
    \item[Signature:]
          A hash of the \gls{tbom}'s contents to verify its integrity and ensure it has not been altered.
\end{description}
This artifact ensures reproducibility, auditability, and cryptographic integrity of the trained model and associated artefacts.

\paragraph{\Gls{ibom} Generation.}

For inference events, the system generates a \gls{ibom} capturing essential details required for auditability.
Specifically, the \gls{ibom} documents raw input features, their encoded representations, model predictions (probabilities for each class), predicted labels, thresholds used in decision-making, and timestamps of the inference event.
Optionally, single-example metrics such as sensitivity and specificity can be computed to provide finer-grained interpretability and transparency.

The \gls{ibom} is crytographically linked to the original \gls{tbom}, establishing a verifiable decision trail from training to each inference.
Similar to the \gls{tbom}, each \gls{ibom} is cryptograhically signed using \gls{dsse} envelopes to guarantee integrity.

\paragraph{\Gls{ibom} Format.}
The documentation of inference-specific information again takes the form of a structured JSON file, supporting the key sections: 

\begin{description}
    \item[Inference Identification:] contains unique inference ID, timestamp, and a hash linking to the specific \gls{tbom} used for this inference session.

    \item[Input Metadata:] captures essential information about the inference input, including input identifier, input dimensions, preprocessing pipeline applied, and any input-specific transformations not covered in the \gls{tbom} training methodology.

    \item[Inference Results:] documents the actual prediction process including:
          \begin{itemize}
              \item \textit{Raw Model Output} contains raw values, intermediate layer activations, and final probability scores before decision thresholding.
              \item \textit{Decision Metrics} provides the final classification decision, confidence score, decision threshold used, distance from threshold, and certainty level assessment.
              \item \textit{Feature Analysis} includes input-specific feature extraction results, concept similarity scores computed for this specific input, and any runtime feature modifications.
          \end{itemize}

    \item[Decision Pathway Tracking:] provides step-by-step documentation of the inference process from input processing through final classification, including any runtime optimizations or modifications applied during inference.

    \item[Temporal Inference Data:] captures inference-specific timing information, computational resources used, and any runtime environmental factors that might affect reproducibility.

    \item[Signature:]
          A hash of the \gls{ibom}'s contents together with the link to \gls{tbom} to verify its integrity and to ensure it has not been altered.

    \item[Runtime Environment:]
          documents hardware, software versions, or computational configurations.
          For models served in a distributed fashion (e.g. via web technology), this would include information about the serving system (which is not the same as the provider system in Figure~\ref{fig:abstract-workflow}).
\end{description}

%% file: sections/06-depend-dbom.tex
\section{Generating \Gls{dbom} Dependably}

Having established the structure and format of the \gls{dbom}, we will now discuss how to generate them dependably---in a way that preserves its integrity and ensures the accountability of the involved parties.

\paragraph{Workflow and Roles.}
To gain a better understanding of the necessary protection mechanisms, we can think about an AI decision as software developers think about a software build pipeline:
In software development, the initial data (the source code) has to be processed (compiled) into an artifact (often a binary).
The artifact, combined with input data, then generates some output.
Similarly, the AI training process transforms the data into an artifact (the model).
This model, combined with some input data, then generates a decision.

Existing work on the security of build pipelines highlights the risk of artifact manipulations.
For \gls{dbom}, this means that we need to protect the model between training and inference from manipulations.
The steps that lead to the decision themselves are protected, as we run them inside a \gls{tee}.

The \gls{tee} also ensures that the training and inference code are integrity protected by attestation.
But for a complete verification, we need a third, safe place where the correct code can be identified.
We solve this problem by identifying different \textit{roles}.
Each member of a role needs to be able to identify their contribution to the pipeline.
We suggest the following roles:
\begin{description}
    \item[Data Owner] provides the training data.
    \item[Model Provider] chooses the training algorithm, trains and provides the model.
    \item[Inference Provider] runs the inference algorithm.
\end{description}

\paragraph{Tamper-Proof Training.}
To protect against interference on an untrusted machine, the training algorithm runs inside a \gls{tee}.
It protects data-in-use, but for the input and output, we need some additional mechanisms.

To protect the input data, the Data Owner either sends it to the Model Provider, who includes it in the initial \gls{tee} state (this is the approach taken for the table-based classification), or uses the SCONE framework to create a SCONE volume that encrypts and authenticates the data (this approach has been used for image-based classification).
When using a SCONE volume, the Data Owner can use a policy to define that only certain Model Provider can get access to the volume.
This prevents the data from being stolen or misused.

To protect the output data, we hash the model and include the hash inside the \gls{tbom}.
Then we sign the model and the \gls{tbom} inside the same \gls{tee} that performs the training.
We achieve this by performing the signing with the same Python script that performs the training.
By signing in the same \gls{tee} that performs the training, we can make sure that every manipulation of the \gls{tbom} or the model will get detected if the attacker is not in posession of the signing key.
It is possible to encrypt the model and the \gls{dbom} before writing it, in case the Training Responsible does not want it to become stolen or public.

Of course, the signature is only as safe as the signing key.
Using the SCONE framework, we can generate the singing key by a trusted third party---the \gls{cas}.
The \gls{cas} ensures that the key is only provided to the training \gls{tee} and cannot be seen from the outside.
The \gls{cas} itself runs inside a \gls{tee} and keeps the signing key secret---only the corresponding verifying key gets published.

The Training Responsible provides the training code and defines which data is accessed and how the \gls{dbom} is created.
They have to identify to \gls{cas} before the generation of the signing key starts or can provide a custom key.
In all cases, the identity of the Training Responsible is tied to the \gls{tbom}, which ensures accountability. In case of a bad configuration, the Training Responsible can be held accountable.

\paragraph{Tamper-Proof Inference.}

Just as the training, the inference is also run inside a \gls{tee}.
It loads the model, if the model has been encrypted, decryption keys are handed out by \gls{cas} only after attestation.
It also reloads the \gls{tbom} and checks the integrity of the model.

The generation of the \gls{ibom} follows the same principles as the \gls{tbom} generation.
The Inference Provider can let \gls{cas} generate a key for signing the \gls{ibom} or provide one.
In both cases, the provider can be held accountable in case of an incorrectly working inference algorithm.

%% file: sections/07-use-case.tex
\begin{figure*}[hbtp]
\centering
    \includegraphics[width=0.81\textwidth]
    {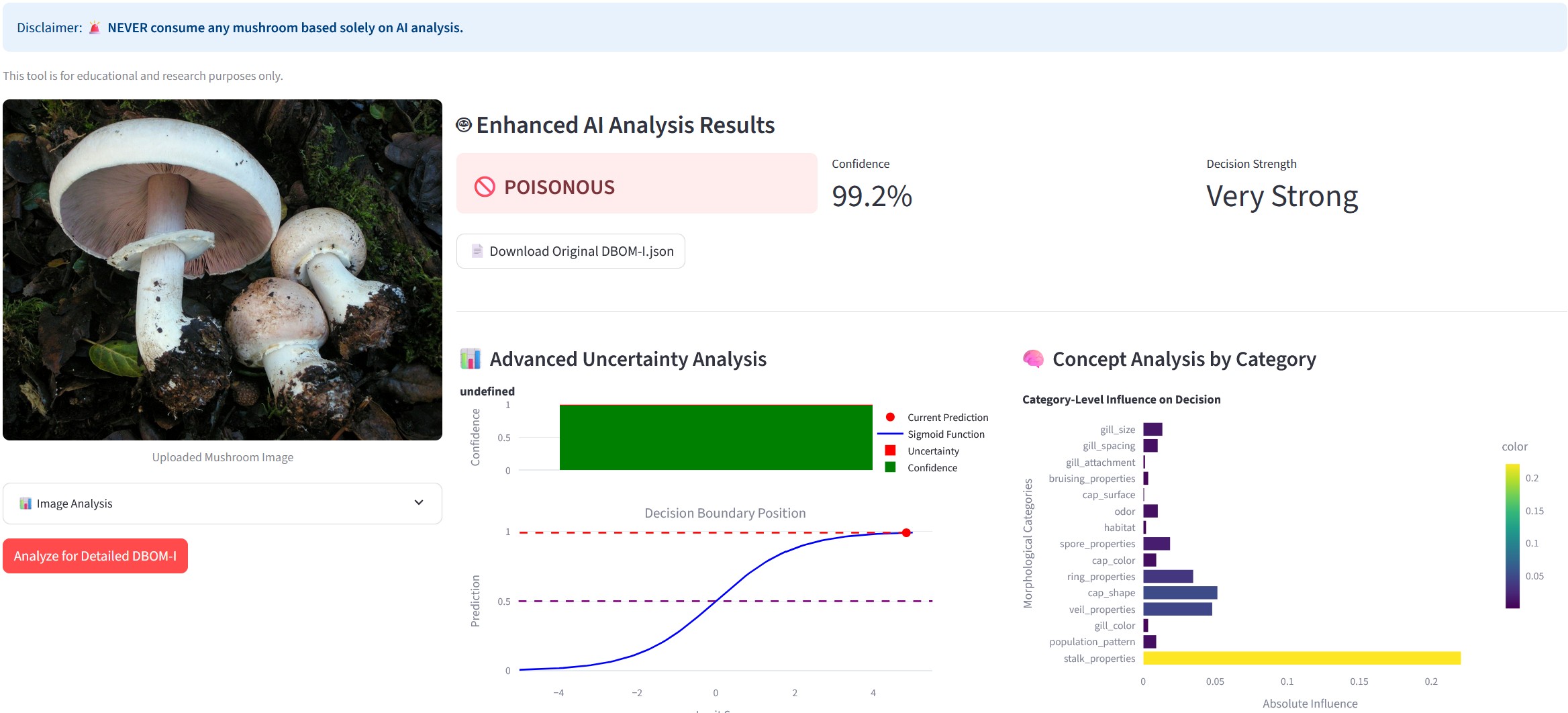}
    \caption{
        \gls{dbom} Inspector -- Showing \fungai's exemplary analysis wrt.\ confidence, uncertainty, and explainability.
    }
    \label{fig:dbom-inspector}
\end{figure*}
\section{Dependable \fungai{}}
\label{sec:use-case}

\fungai{} is an AI-based image classification application designed to determine, from an image of a mushroom, whether that mushroom is poisonous or edible. The entire AI system is implemented with dependable execution and integrated with \gls{dbom} support. The format and generation process of the \gls{dbom} follow the description provided in the previous section. It includes all detailed information related to the generic AI system outlined below.

\paragraph{\wfiterm{image}{Data}.}
We use the Mushrooms dataset\footnote{\url{https://www.kaggle.com/datasets/derekkunowilliams/mushrooms}}, comprising 8,468 images labeled as edible, conditionally edible, poisonous, or deadly. For binary classification, we merge edible and conditionally edible into the \emph{edible} class, and poisonous and deadly into the \emph{poisonous} class, resulting in 2,895 edible and 5,573 poisonous samples.

\paragraph{\wfiterm{wave-square}{Model}.}
Given the limited data, we adopt a model suitable for few-shot learning with built-in interpretability. Our binary classifier is based on the Ph-CBM architecture~\cite{yuksekgonul2022post}, utilizing pretrained multimodal backbone CLIP (ViT-L/14)\footnote{\url{https://github.com/openai/CLIP}}. The concept set is derived from the tabular Mushroom Classification dataset\footnote{\url{https://www.kaggle.com/datasets/uciml/mushroom-classification}}, consisting of attribute–value pairs (e.g., cap color: red).
We keep the CLIP backbone frozen and add four fully connected layers as the final binary classifier. For \fungai{}, we fine-tune only the hyperparameters of these added layers.

\paragraph{\wfiterm{chart-diagram}{Process}.}
We split the dataset into $80\%$ for training and $20\%$ for testing. During training, we apply 5-fold stratified cross-validation for balanced class representation across folds. The model’s performance is then evaluated on the held-out testing set.

\paragraph{\wfiterm{microchip}{Provider System}.}
We use Intel SGX in combination with SCONE to protect the training and inference process.
Training and inference are run in separate containers and can be run on different machines.
The \gls{dbom} is cryptographically signed with a key generated by \gls{cas}, the verifying key can be retrieved from there.
The signing key is never shown to any involved party.
The DBOM includes hashes of the model and training data, thus preserving their integrity.
For now, we do not encrypt the model after training, but we might add this at a later point in time.

\paragraph{\wfiterm{server}{Deployer System}.}
The trained classification model is deployed on a server with Intel SGX support, while the user interface operates on the consumer’s device (tested on a personal computer).
For now, we always used the same server for training and inference, but this is optional.
Consumers submit input data (e.g., mushroom images) via the interface, which interacts with the server to perform inference. Inference is conducted within a \gls{tee}, ensuring secure and trustworthy execution.

\paragraph{\wfiterm{gavel}{Decision}.}
In addition to displaying the model’s final prediction, we also present detailed information, including concept contributions, showing the weighted influence of each concept on the decision, and the output's confidence score (i.e., predicted probability).


%% file: sections/07-eval.tex
\section{Evaluation \& Discussion}

The generation of a tamper-proof \gls{dbom} does come with additional costs.
Hardware support is in general not a cost factor, as all modern Intel processors for server systems come with \gls{sgx} extensions. However, the generation itself suffers from additional overhead because
\begin{enumerate*}[label=\alph*)]
    \item extra bookkeeping is needed at many distinct steps in the AI decision system and
    \item the training process and any other computation is executed in a tamper-proof, yet performance-constrained system.
    For example, enclaves in \gls{sgx} have a limited cache and every system call needs to be handled in a specific way to avoid data leakage
\end{enumerate*}.
It should be noted, though, that these steps are only needed for to-be-deployed AI models.
During development and optimisation of a concrete training pipeline, there is no need for tamper-proof, yet slow, execution.  
Instead, this can be confined to model versions that are to be deployed, for which then a matching \gls{dbom} is produced. 

To get an idea of the performance penalty introduced by confidential computing, we evaluated the concept-based (with concept features stored in CSV) variant of \fungai{}.
We did not evaluate the full, image and concept-based training because we so far cannot use GPU-based training, as confidential GPUs are just entering the market and are not (yet) available to us.
Hence, image-based algorithms would need to run on a confidential CPU, which cannot compete with GPU-based training.
Furthermore the many filesystem interactions induced  by the high number of image files caused issues along the cryptography pipeline. We might address this with unencrypted images (we only need to protect integrity). Therefore, our experiments for now focus on the implemented principles of \gls{dbom} generation and its feasibility.

We used a Intel Xeon Silver 4314 CPU from the Icelake generation, which has 16 physical and 32 logical cores.
We ran Ubuntu 24.04.01 on the host system and Alpine 3.22 inside the containers.
The later choice was motivated by a missing adaptation of the newest glibc version by SCONE, which we needed for the training libraries.
Since we cannot fork subprocesses inside enclaves\footnote{There is ongoing work to lift this restriction, but we are not aware of any solution that would have helped us.}, we used the new, experimental free-threading Python, introduced in version 3.13.

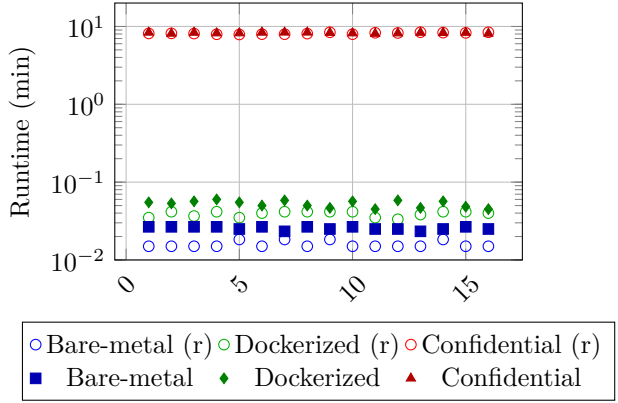
\begin{figure}[h]
    \centering
    \begin{tikzpicture}
\begin{axis}[
    width=7cm,
    height=5cm,
    xlabel={},
    ylabel={Runtime (min)},
    ymode=log,
    log basis y=10,
    ymin=0.01,
    ymax=20,
    legend style={at={(0.5,-0.24)}, anchor=north, legend columns=3},
    grid=major,
    xticklabel style={rotate=45},
]

\addplot[mark=o, color=blue, only marks] 
table[
        col sep=comma,
    x=x,
    y=Noconf-rampup,
] {figures/results_converted.csv};
\addlegendentry{Bare-metal (r)}

\addplot[mark=o, color=green!70!black, only marks] table[
        col sep=comma,
    x=x,
    y=Docker-rampup,
] {figures/results_converted.csv};
\addlegendentry{Dockerized (r)}

\addplot[mark=o, color=red, only marks] table[
        col sep=comma,
    x=x,
    y=Conf-rampup,
] {figures/results_converted.csv};
\addlegendentry{Confidential (r)}

\addplot[mark=square*, color=blue!70!black, only marks] table[
        col sep=comma,
    x=x,
    y=Noconf-full,
] {figures/results_converted.csv};
\addlegendentry{Bare-metal}

\addplot[mark=diamond*, color=green!50!black, only marks] table[
        col sep=comma,
    x=x,
    y=Docker-full,
] {figures/results_converted.csv};
\addlegendentry{Dockerized}

\addplot[mark=triangle*, color=red!70!black, only marks] table[
        col sep=comma,
    x=x,
    y=Conf-full,
] {figures/results_converted.csv};
\addlegendentry{Confidential}

\end{axis}
\end{tikzpicture}
    \caption{Runtimes for different configurations. The (r) variants refer to the runs where only dependencies were loaded and no training took place.}
    \label{fig:placeholder}
\end{figure}

In our preliminary experiments, the results of which are shown in Figure \ref{fig:placeholder}, we could observe a runtime increase by a factor of 157 compared to a container-based non-confidential variant.
Compared to a bare-metal variant, we even saw an increase by a factor of 325.
However, we could observe that almost 99\% of the runtime of a confidential training process is needed only for the setup of our Python script~(starting the interpreter, importing dependencies, loading data, etc.).

The percentage likely decreases with more complex training pipelines, since the main source of performance slowdown is the much slower loading of dependencies.
In SCONE, dependencies also get encrypted and signed, which means that they have to be decrypted if the script starts.
Moreover, the necessary filesystem interactions are slowed down by the need to leave and re-enter the enclave for each one of them.

Summarizing our first empirical results, we see a considerable slowdown that might confine the tamper-proof \gls{dbom} approach to only the deployment versions of high-stake AI applications. Of course, \glspl{dbom} are still useful even if not made tamper-proof, since they provide a record of all relevant information. Furthermore, we expect that the \gls{dbom} solution will become more attractive with confidential GPUs becoming more mature and cheaper, together with further improvements in the implementation of Confidential Computing.  Moreover, we can think of training services that need to load the libraries only once thus reduce the additional time needed.

%% file: sections/08-ecosystem.tex
\section{Towards a \Gls{dbom} Ecosystem}

Similar to how the advent of SBOMs has led to an ecosystem of tools, we can conceive different tools that use \glspl{dbom}, each contributing to the dependability of high-stake AI decision.
We here propose a small, incomplete, set of tools we imagine to leverage \glspl{dbom}.

\paragraph{\faIcon{magnifying-glass-plus} \gls{dbom} Inspector:}
\gls{dbom} facilitates visualization and inspection of the training and inference processes.
To demonstrate this, we have already implemented an exemplary interface  highlighting key capabilities~(cf. Figure~\ref{fig:dbom-inspector}).\footnote{\url{https://huggingface.co/spaces/fungi00/}}
During training, the system visualizes essential data insights, such as the training accuracy per epoch, together with all learned concepts along with their corresponding importance scores.
For inference, given an input image, the interface retrieves concept correlation information from \gls{ibom}.
Users can interactively modify concept correlations, enabling real-time intervention to observe how such changes affect the final prediction.


\paragraph{\faIcon{fingerprint} Integrity Checker:}
As the \gls{dbom} is just a format, there will be instances of \glspl{dbom} that violate certain  aspects.
A check whether a file is a valid instance of \gls{dbom} (i.e.\ adheres to the proper the right structure) seems worthwhile to have, together with a check that the \gls{dbom} has integrity--- the entire cryptographic information~(mostly related to confidential computing) is intact.
Hence, a first step in many \gls{dbom} related pipelines could be to run a integrity checker on a \gls{dbom} and react accordingly, in case it is invalid or damaged.

\paragraph{\faIcon{section}~Compliance Checker:}
One level higher in the abstraction hierarchy is the check for compliance.
In the future, we can envision standards (e.g. by ISO or CENELEC) to mandate certain aspects of AI training.
As a placative example, one could prescribe that testing accuracy must be at least 95\%.
In consequence, an auditing authority could use the \gls{tbom} to check if the corresponding model is non-compliant.

\paragraph{\faIcon{binoculars} Vigilance Checker:}
Taking inspiration from the medical domain, where vigilance systems track medical products in the field, similar systems can be established for AI.
\gls{dbom}-producing applications could be forced to occasionally report \glspl{dbom} to a central authority.
The authority would then check the information contained and act accordingly, possibly considering the temporal evolution across several \glspl{dbom}.  

%% file: sections/09-conclusion.tex
\section{Conclusion}
\label{sec:conclusion}
This paper is about breathing life into the \gls{dbom} concept, presenting the first ever functional workflow that supports the creation of tamper-proof, verifiable, and complete traces of AI decisions. We are working on the various ecosystem components with enthusiasm, as well as on further optimisations with respect to the cryptograhic overhead incurred.

%% file: sections/11-supp.tex
\clearpage

\title{Supplementary material for \\
``A Workflow for Full Traceability of AI Decision''}

\maketitle

\setcounter{page}{1}

\appendix

\renewcommand{\theequation}{A\arabic{equation}}
\renewcommand{\thetable}{A\arabic{table}}
\renewcommand{\thefigure}{A\arabic{figure}}


\section*{Introduction}

In the supplementary material, we first provide an in-depth description of the implementation details of our \gls{dbom} inspector. Subsequently, we include additional screenshots to illustrate its user interface and operational features. Finally, we present the raw datasets corresponding to both the \gls{tbom} and \gls{ibom} for completeness and reproducibility.

\section{Appendix A: \gls{dbom} Inspector}

\paragraph{Implementation Details.}

We implemented the interface using the \emph{Hugging Face Spaces} platform. To deploy our demo, we created an anonymous GitHub account using a temporary email service (\emph{ProtonMail}) to maintain privacy. The application was then hosted on the \emph{Hugging Face Spaces} platform, which provides free public deployment for machine learning demos. Ultimately, the final version of our project is accessible at \url{https://huggingface.co/spaces/fungi00/fungiclassifier}.

\begin{figure}
    \centering
    \includegraphics[width=\linewidth]{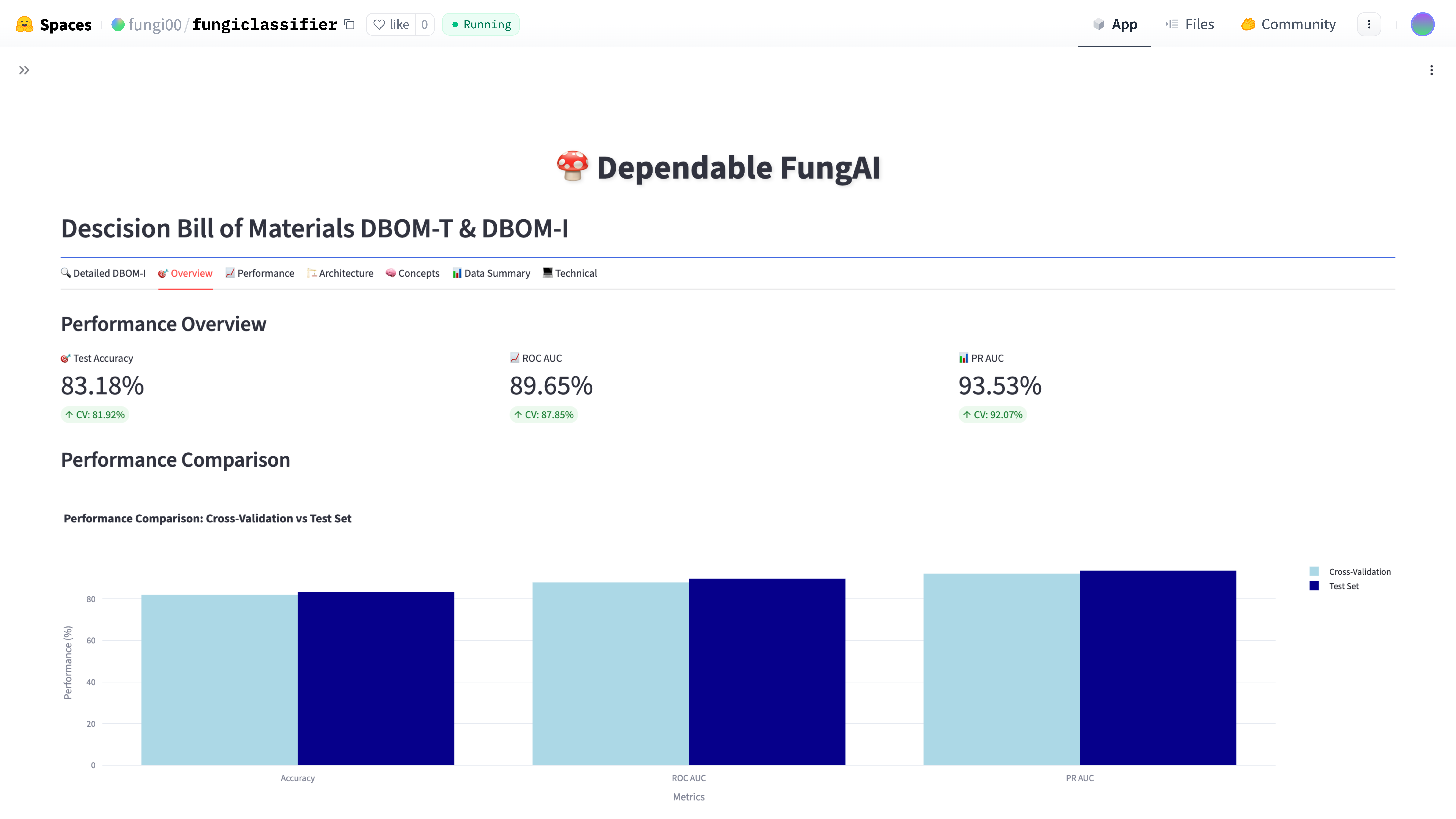}
    \caption{Overview Tab}
    \label{fig:inspector-overview}
\end{figure}

\paragraph{Inspector Interface.} As described in the main paper, our inspector is designed to visualize both the \gls{tbom} and \gls{ibom}. The Overview tab presents the model's accuracy on the test set, along with the corresponding ROC AUC and PR AUC scores. A bar plot comparing these performance metrics between the validation and test sets is shown in Figure~\ref{fig:inspector-overview}.

\begin{figure}
    \centering
    \includegraphics[width=\linewidth]{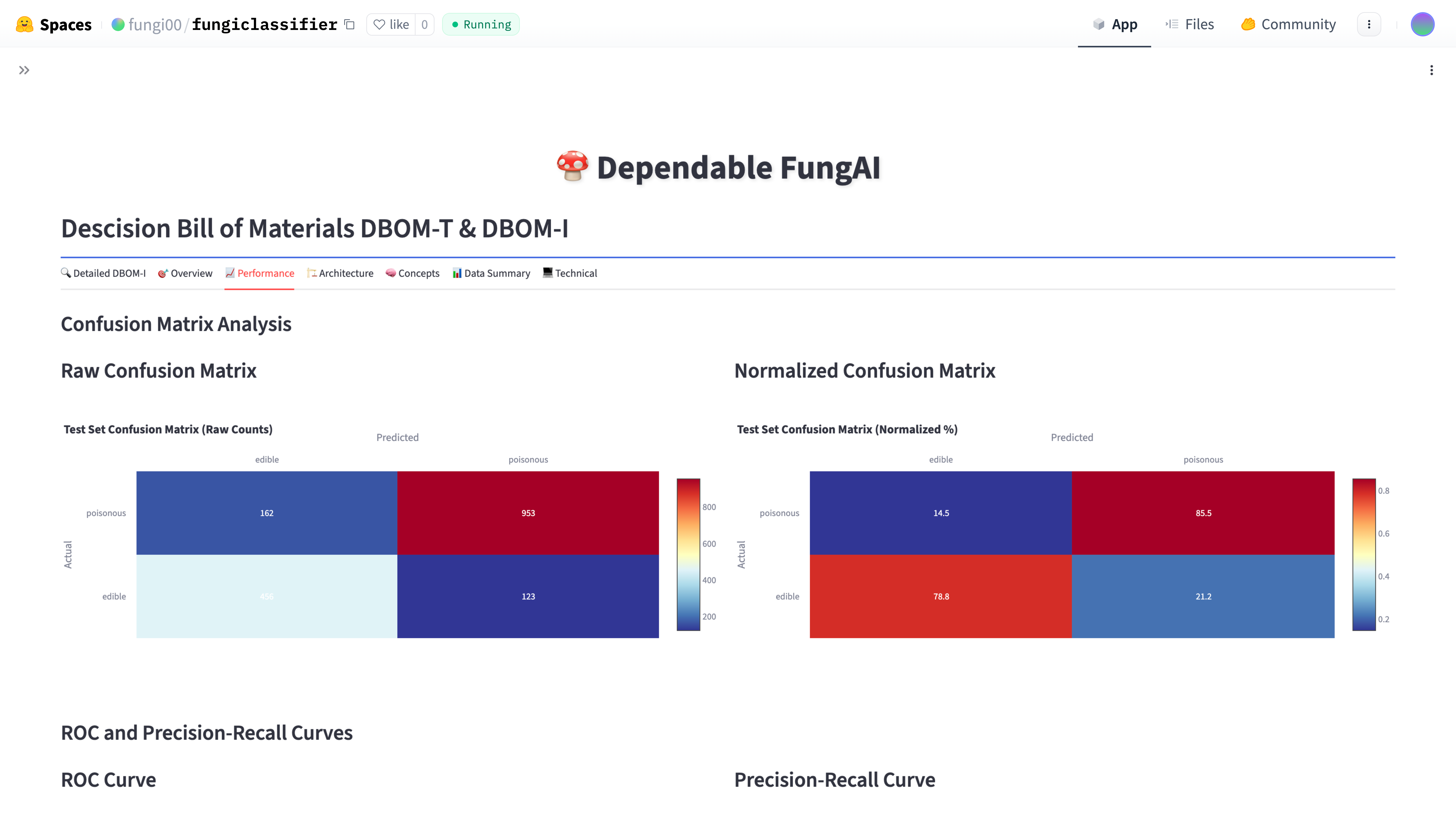}
    \caption{Performance Tab}
    \label{fig:performance_tab}
\end{figure}

In the Performance tab, we provide a detailed analysis of the test set via a confusion matrix, along with the ROC and precision-recall curves. Additionally, we visualize the evolution of these performance metrics throughout the training process, as illustrated in Figure~\ref{fig:performance_tab}.

\begin{figure}
    \centering
    \includegraphics[width=\linewidth]{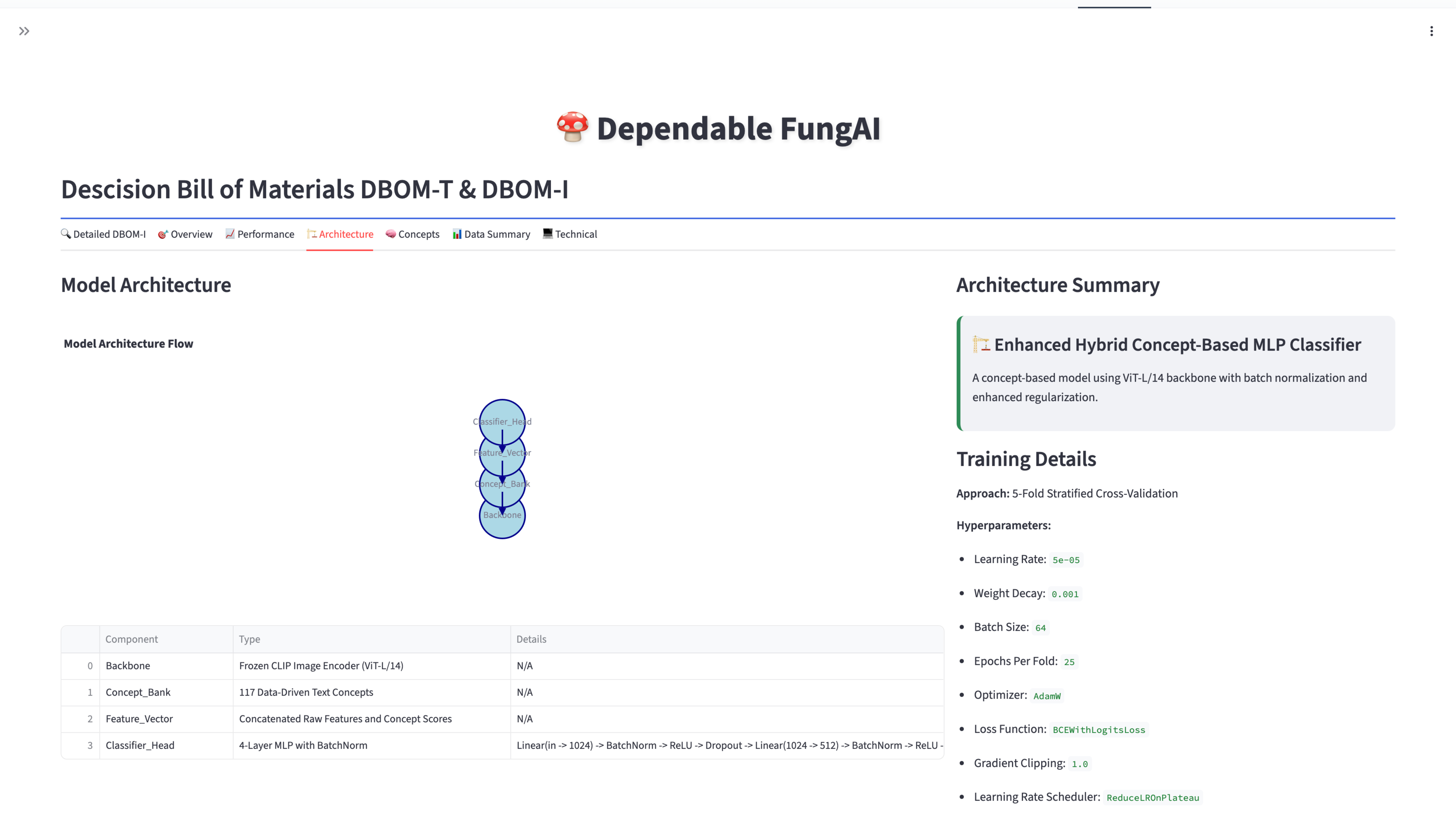}
    \caption{Architecture Overview}
    \label{fig:inspector-architecture}
\end{figure}

\begin{figure}
    \centering
    \includegraphics[width=\linewidth]{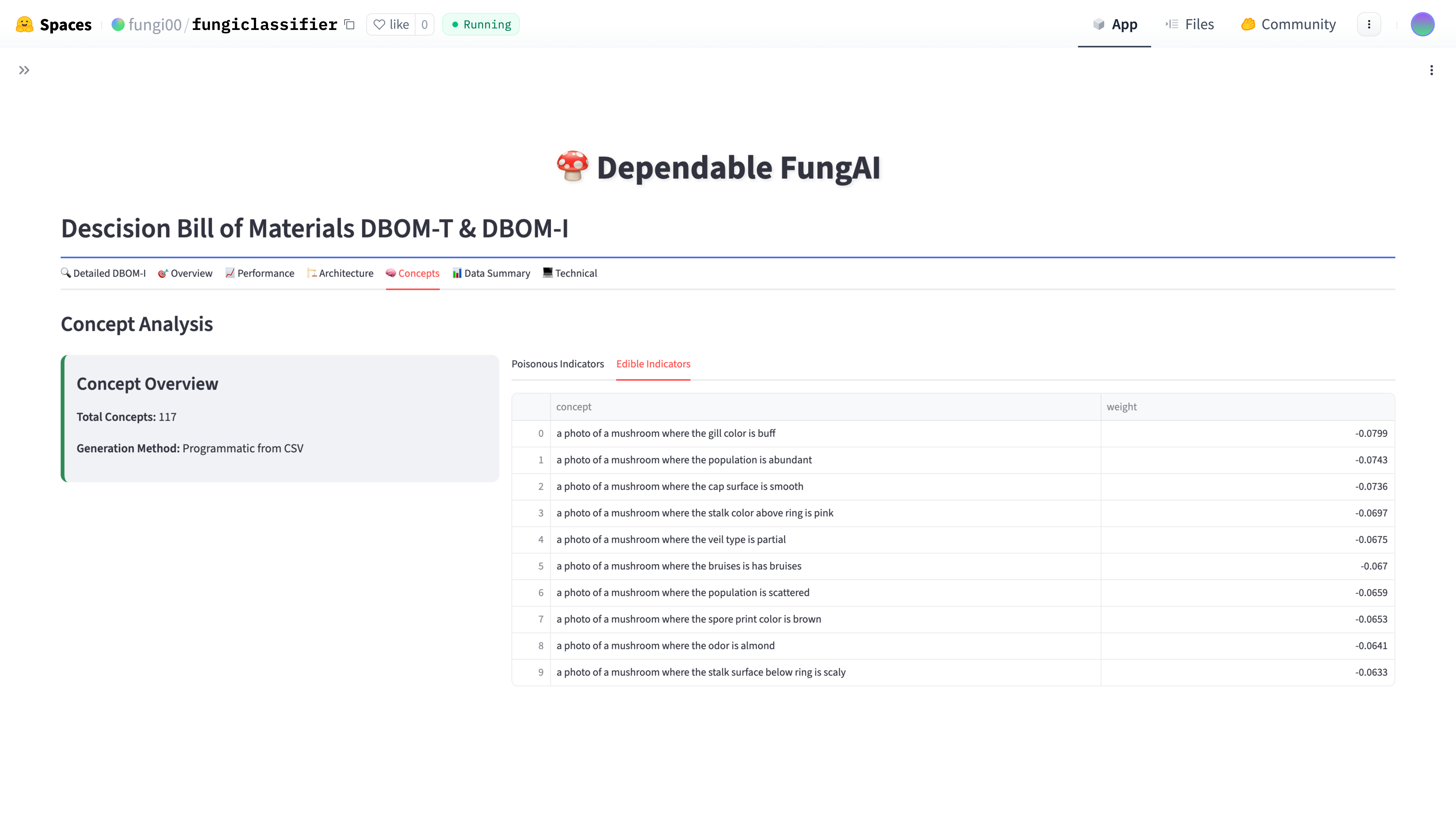}
    \caption{Concept Tab Overview}
    \label{fig:inspector-concept}
\end{figure}

The Architecture tab summarizes the model architecture and lists the key hyperparameters used during training (Figure~\ref{fig:inspector-architecture}). In the Concepts tab, we provide a conceptual summary, highlighting the most influential concepts for both poisonous and edible predictions (Figure~\ref{fig:inspector-concept}).

\begin{figure}
    \centering
    \includegraphics[width=\linewidth]{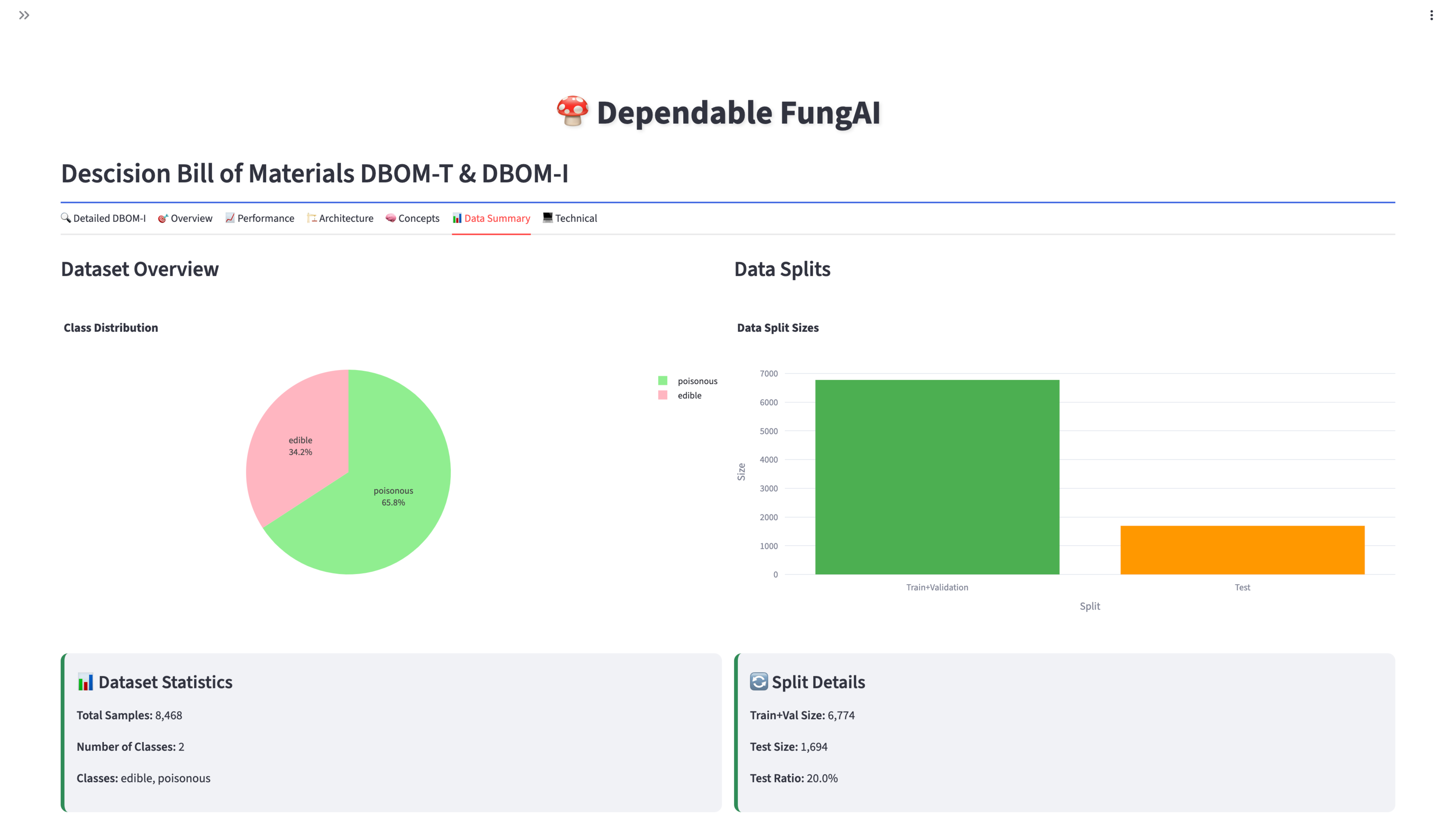}
    \caption{Data Summary Overview}
    \label{fig:inspector-dataset}
\end{figure}

\begin{figure}
    \centering
    \includegraphics[width=\linewidth]{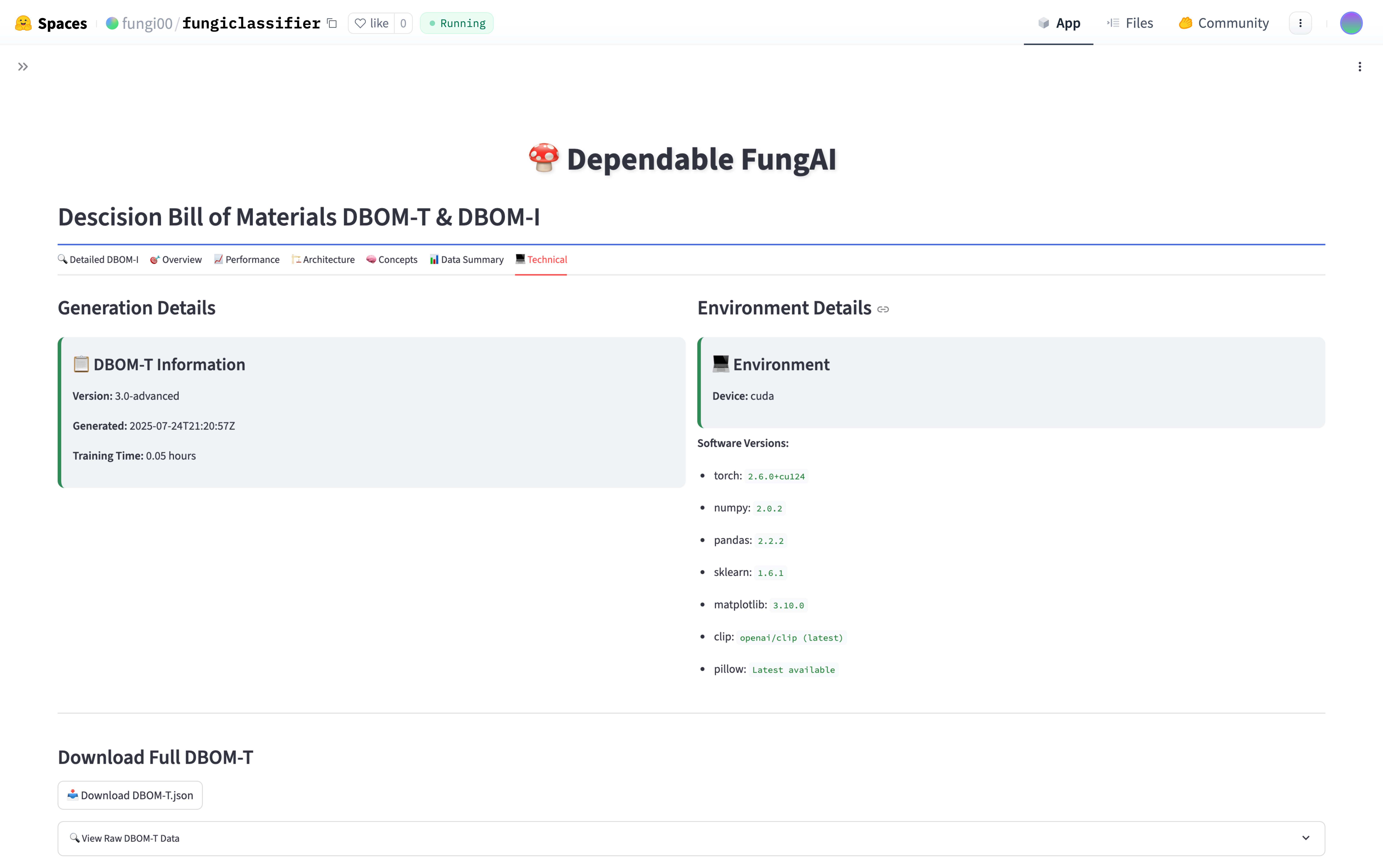}
    \caption{Technical Details Overview}
    \label{fig:inspector-technical}
\end{figure}

A dataset overview is provided in Figure~\ref{fig:inspector-dataset}, and technical details of the environment setup are documented in the Technical tab, as shown in Figure~\ref{fig:inspector-technical}.

\begin{figure}
    \centering
    \includegraphics[width=\linewidth]{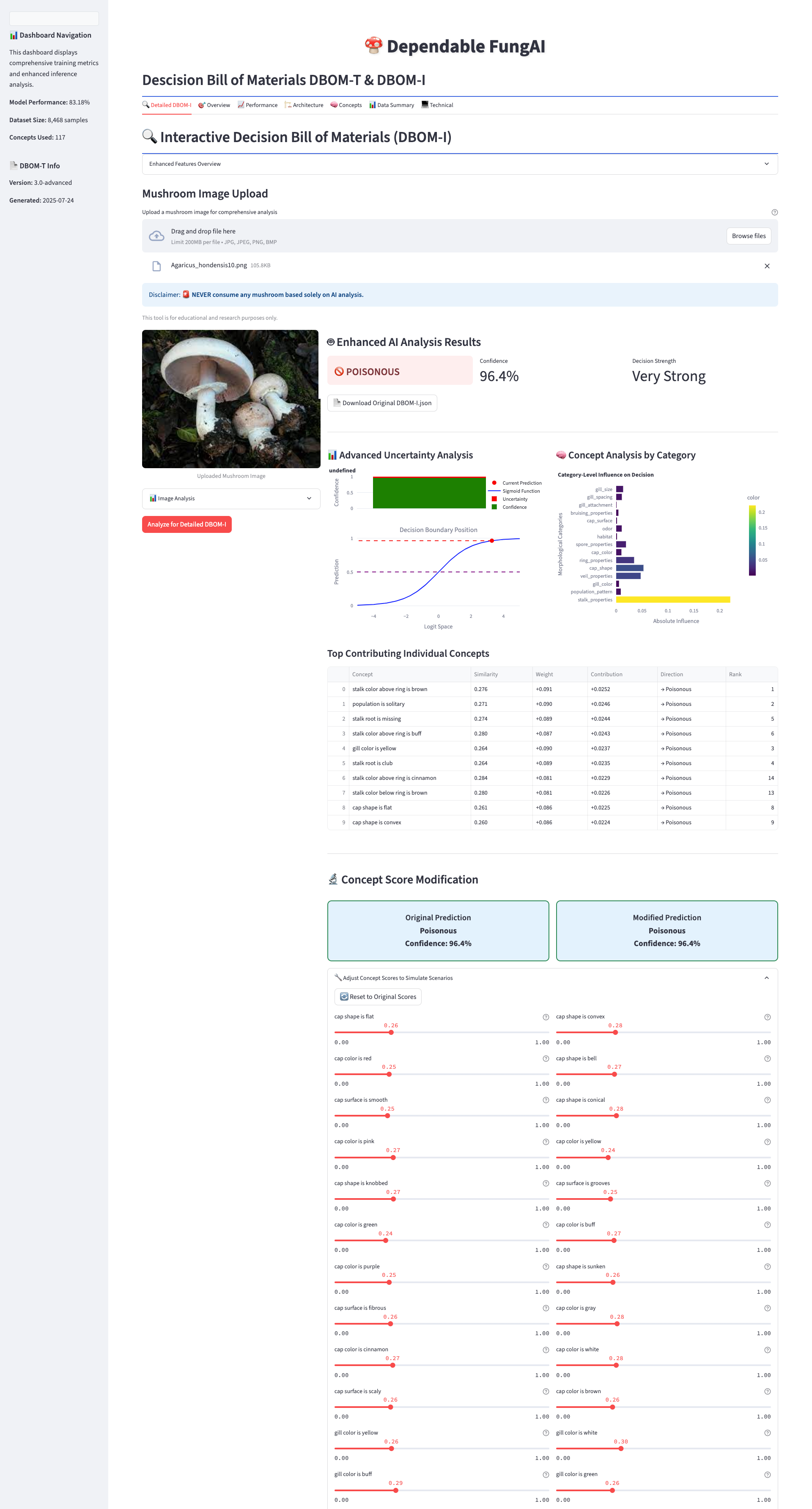}
    \caption{Interactive Interface}
    \label{fig:inspector-ibom}
\end{figure}

We also implemented an interactive interface, as shown in Figure~\ref{fig:inspector-ibom}, for the \gls{ibom}, where users can upload their own images. Upon clicking "Analyze," the model's prediction and confidence score are displayed, along with the related concept activations and uncertainty analysis. Furthermore, users can modify the relevance scores of specific concepts to observe how such changes affect the final prediction.

\section{Appendix B: Full JSON Files of \gls{dbom}}

The JSON files of \gls{dbom} contain both the \gls{tbom} and \gls{ibom} components. As described in the main paper, the \gls{tbom} encompasses comprehensive information about the dataset, model architecture, and training process. The raw content of the \gls{tbom} is provided in the file \texttt{DBOM-T.json}. In contrast, the \gls{ibom} contains instance-specific information for a given image. An example of an \gls{ibom} corresponding to the image shown in Figure~\ref{fig:inspector-ibom} is provided, and its raw data can be found in the file \texttt{DBOM-I.json}.